\documentclass[journal]{IEEEtran}

\usepackage[numbers]{natbib}
\providecommand{\tightlist}{%
  \setlength{\itemsep}{0pt}\setlength{\parskip}{0pt}}

\usepackage[hidelinks,unicode=true]{hyperref}
\usepackage{amssymb,amsmath}

\DeclareMathOperator*{\argmax}{argmax}
\usepackage{graphicx}

\newcommand{\subparagraph}{}
\usepackage{titlesec}
\titlespacing{\section}{0pt}{*0.5}{*0.3}
\titlespacing{\subsection}{0pt}{*0.5}{*0.3}

\begin{document}
\title{A Brief Survey of Deep Reinforcement Learning}

\author{Kai Arulkumaran, Marc Peter Deisenroth, Miles Brundage, Anil Anthony Bharath\vspace{-1.5em}}

%
%

\markboth{IEEE Signal Processing Magazine, Special Issue on Deep Learning for Image Understanding (arXiv extended version)}%
{}

\maketitle

\begin{abstract}
Deep reinforcement learning is poised to revolutionise the field of AI
and represents a step towards building autonomous systems with a higher
level understanding of the visual world. Currently, deep learning is
enabling reinforcement learning to scale to problems that were
previously intractable, such as learning to play video games directly
from pixels. Deep reinforcement learning algorithms are also applied to
robotics, allowing control policies for robots to be learned directly
from camera inputs in the real world. In this survey, we begin with an
introduction to the general field of reinforcement learning, then
progress to the main streams of value-based and policy-based methods.
Our survey will cover central algorithms in deep reinforcement learning,
including the deep \(Q\)-network, trust region policy optimisation, and
asynchronous advantage actor-critic. In parallel, we highlight the
unique advantages of deep neural networks, focusing on visual
understanding via reinforcement learning. To conclude, we describe
several current areas of research within the field.
\end{abstract}

\section{Introduction}\label{introduction}

One of the primary goals of the field of artificial intelligence (AI) is
to produce fully autonomous agents that interact with their environments
to learn optimal behaviours, improving over time through trial and
error. Crafting AI systems that are responsive and can effectively learn
has been a long-standing challenge, ranging from robots, which can sense
and react to the world around them, to purely software-based agents,
which can interact with natural language and multimedia. A principled
mathematical framework for experience-driven autonomous learning is
reinforcement learning (RL) \citep{sutton1998reinforcement}. Although RL
had some successes in the past
\citep{tesauro1995temporal, singh2002optimizing, kohl2004policy, ng2006autonomous},
previous approaches lacked scalablity and were inherently limited to
fairly low-dimensional problems. These limitations exist because RL
algorithms share the same complexity issues as other algorithms: memory
complexity, computational complexity, and in the case of machine
learning algorithms, sample complexity \citep{strehl2006pac}. What we
have witnessed in recent years---the rise of deep learning, relying on
the powerful \emph{function approximation} and \emph{representation
learning} properties of deep neural networks---has provided us with new
tools to overcoming these problems.

The advent of deep learning has had a significant impact on many areas
in machine learning, dramatically improving the state-of-the-art in
tasks such as object detection, speech recognition, and language
translation \citep{lecun2015deep}. The most important property of deep
learning is that deep neural networks can automatically find compact
low-dimensional representations (features) of high-dimensional data
(e.g., images, text and audio). Through crafting inductive biases into
neural network architectures, particularly that of hierarchical
representations, machine learning practitioners have made effective
progress in addressing the curse of dimensionality
\citep{bengio2013representation}. Deep learning has similarly
accelerated progress in RL, with the use of deep learning algorithms
within RL defining the field of ``deep reinforcement learning'' (DRL).
The aim of this survey is to cover both seminal and recent developments
in DRL, conveying the innovative ways in which neural networks can be
used to bring us closer towards developing autonomous agents. For a more
comprehensive survey of recent efforts in DRL, including applications of
DRL to areas such as natural language processing
\citep{ranzato2016sequence, bahdanau2017actor}, we refer readers to the
overview by Li \citep{li2017deep}.

Deep learning enables RL to scale to decision-making problems that were
previously intractable, i.e., settings with high-dimensional state and
action spaces. Amongst recent work in the field of DRL, there have been
two outstanding success stories. The first, kickstarting the revolution
in DRL, was the development of an algorithm that could learn to play a
range of Atari 2600 video games at a superhuman level, directly from
image pixels \citep{mnih2015human}. Providing solutions for the
instability of function approximation techniques in RL, this work was
the first to convincingly demonstrate that RL agents could be trained on
raw, high-dimensional observations, solely based on a reward signal. The
second standout success was the development of a hybrid DRL system,
AlphaGo, that defeated a human world champion in Go
\citep{silver2016mastering}, paralleling the historic achievement of
IBM's Deep Blue in chess two decades earlier \citep{campbell2002deep}
and IBM's Watson DeepQA system that beat the best human Jeopardy!
players \citep{ferrucci2010building}. Unlike the handcrafted rules that
have dominated chess-playing systems, AlphaGo was composed of neural
networks that were trained using supervised and reinforcement learning,
in combination with a traditional heuristic search algorithm.

\begin{figure*}
  \centering
  \includegraphics[width=0.85\textwidth]{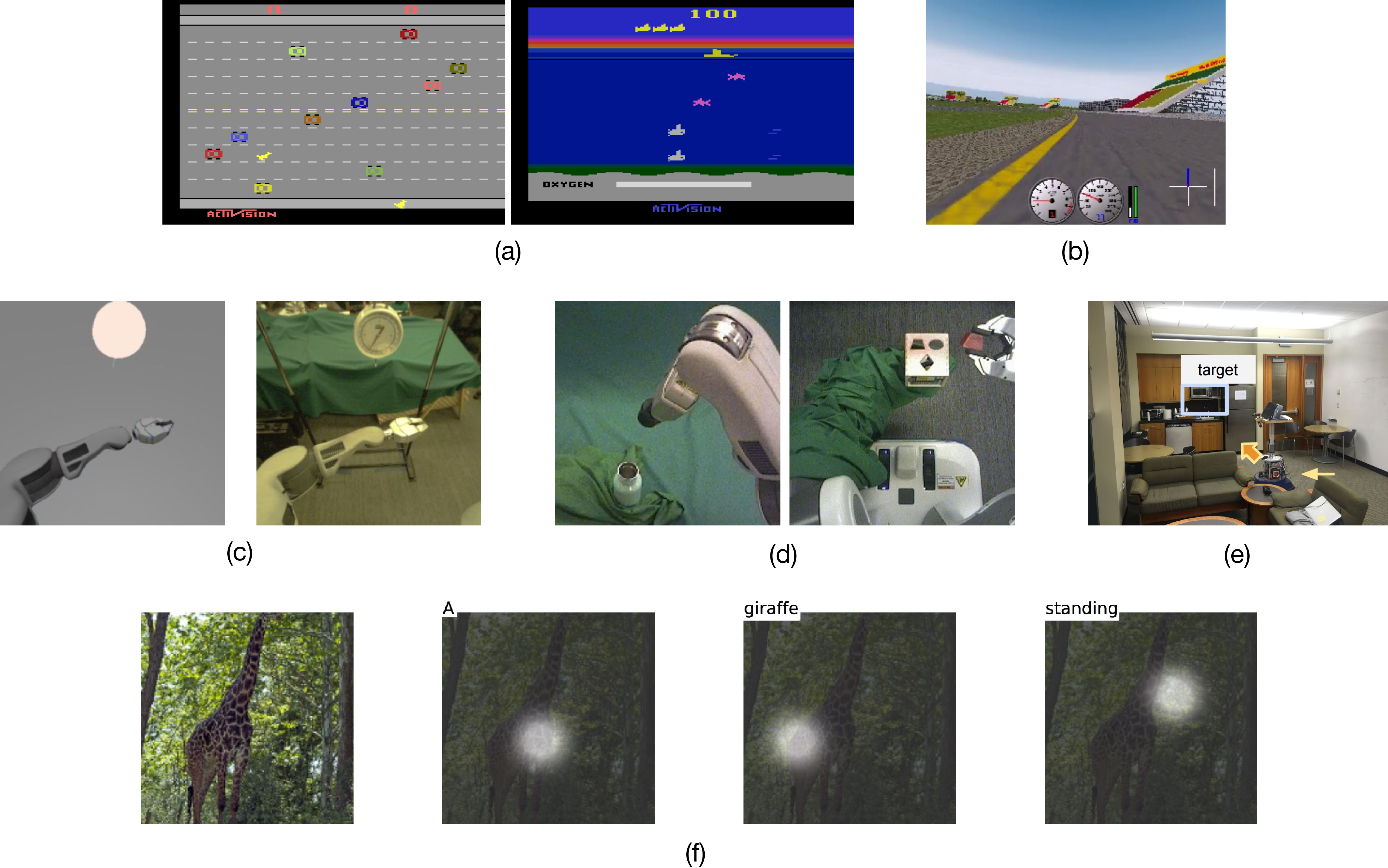}
  \caption{A range of visual RL domains. \textbf{(a)} Two classic Atari 2600 video games, ``Freeway'' and ``Seaquest'', from the Arcade Learning Environment (ALE) \cite{bellemare2015arcade}. Due to the range of supported games that vary in genre, visuals and difficulty, the ALE has become a standard testbed for DRL algorithms \cite{mnih2015human,oh2015action,hausknecht2015deep,schulman2015trust,stadie2015incentivizing,wang2016dueling,mnih2016asynchronous}. As we will discuss later, the ALE is one of several benchmarks that are now being used to standardise evaluation in RL. \textbf{(b)} The TORCS car racing simulator, which has been used to test DRL algorithms that can output continuous actions \cite{koutnik2013evolving,lillicrap2016continuous,mnih2016asynchronous} (as the games from the ALE only support discrete actions). \textbf{(c)} Utilising the potentially unlimited amount of training data that can be amassed in robotic simulators, several methods aim to transfer knowledge from the simulator to the real world \cite{christiano2016transfer,rusu2017sim,tzeng2016towards}. \textbf{(d)} Two of the four robotic tasks designed by Levine et al. \cite{levine2016end}: screwing on a bottle cap and placing a shaped block in the correct hole. Levine et al. \cite{levine2016end} were able to train visuomotor policies in an end-to-end fashion, showing that visual servoing could be learned directly from raw camera inputs by using deep neural networks. \textbf{(e)} A real room, in which a wheeled robot trained to navigate the building is given a visual cue as input, and must find the corresponding location \cite{zhu2017target}. \textbf{(f)} A natural image being captioned by a neural network that uses reinforcement learning to choose where to look \cite{xu2015show}. By processing a small portion of the image for every word generated, the network can focus its attention on the most salient points. Figures reproduced from \cite{bellemare2015arcade,lillicrap2016continuous,tzeng2016towards,levine2016end,zhu2017target,xu2015show}, respectively.}
  \label{fig:examples}
\end{figure*}

DRL algorithms have already been applied to a wide range of problems,
such as robotics, where control policies for robots can now be learned
directly from camera inputs in the real world
\citep{levine2016end, levine2016learning}, succeeding controllers that
used to be hand-engineered or learned from low-dimensional features of
the robot's state. In a step towards even more capable agents, DRL has
been used to create agents that can meta-learn (``learn to learn'')
\citep{duan2016rl, wang2017learning}, allowing them to generalise to
complex visual environments they have never seen before
\citep{duan2016rl}. In Figure \ref{fig:examples}, we showcase just some
of the domains that DRL has been applied to, ranging from playing video
games \citep{mnih2015human} to indoor navigation \citep{zhu2017target}.

Video games may be an interesting challenge, but learning how to play
them is not the end goal of DRL. One of the driving forces behind DRL is
the vision of creating systems that are capable of learning how to adapt
in the real world. From managing power consumption
\citep{tesauro2008managing} to picking and stowing objects
\citep{levine2016learning}, DRL stands to increase the amount of
physical tasks that can be automated by learning. However, DRL does not
stop there, as RL is a general way of approaching optimisation problems
by trial and error. From designing state-of-the-art machine translation
models \citep{zoph2017neural} to constructing new optimisation functions
\citep{li2017learning}, DRL has already been used to approach all manner
of machine learning tasks. And, in the same way that deep learning has
been utilised across many branches of machine learning, it seems likely
that in the future, DRL will be an important component in constructing
general AI systems \citep{lake2016building}.

\section{Reward-driven Behaviour}\label{reward-driven-behaviour}

\begin{figure*}[ht]
  \centering
  \includegraphics[width=0.75\textwidth]{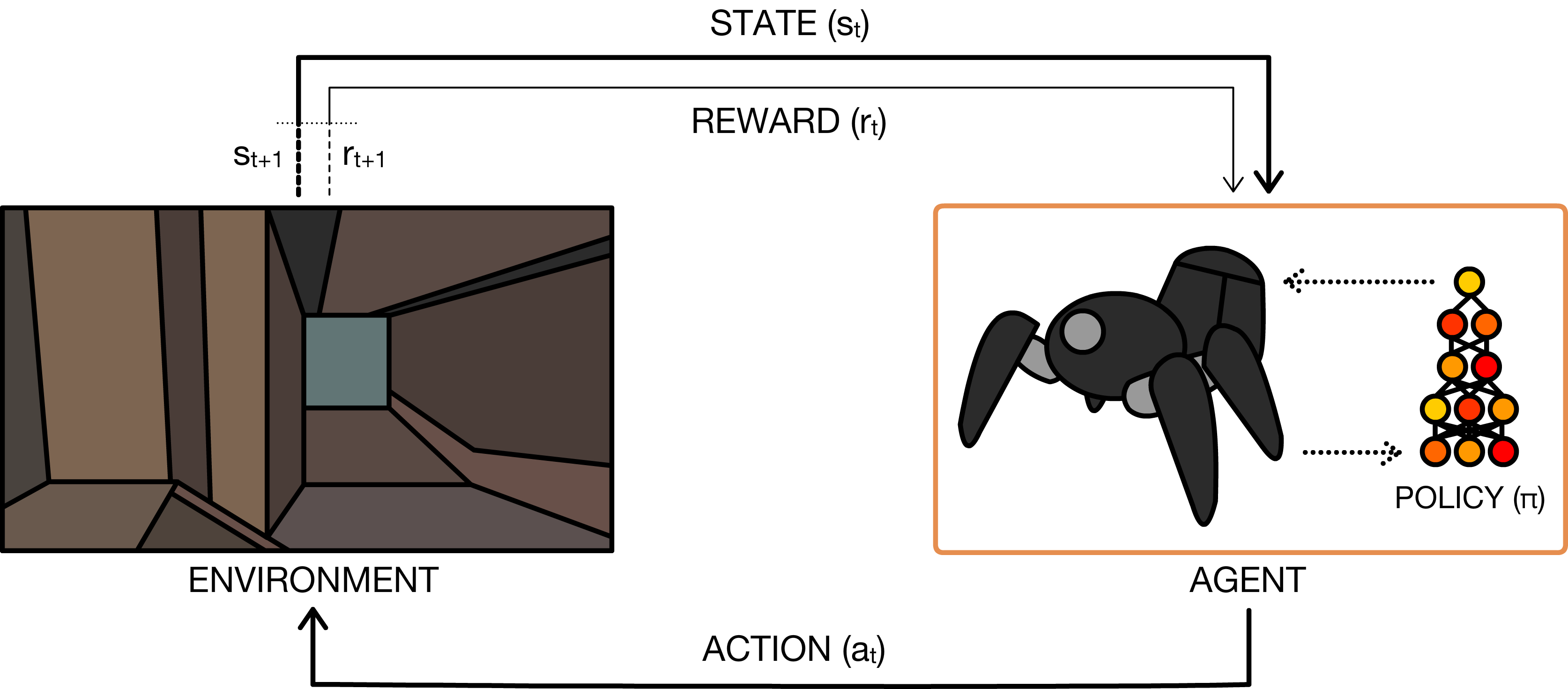}
  \caption{The perception-action-learning loop. At time $t$, the agent receives state $\mathbf{s}_t$ from the environment. The agent uses its policy to choose an action $\mathbf{a}_t$. Once the action is executed, the environment transitions a step, providing the next state $\mathbf{s}_{t+1}$ as well as feedback in the form of a reward $r_{t+1}$. The agent uses knowledge of state transitions, of the form $(\mathbf{s}_t, \mathbf{a}_t, \mathbf{s}_{t+1}, r_{t+1})$, in order to learn and improve its policy.}
  \label{fig:pal}
\end{figure*}

Before examining the contributions of deep neural networks to RL, we
will introduce the field of RL in general. The essence of RL is learning
through \emph{interaction}. An RL agent interacts with its environment
and, upon observing the consequences of its actions, can learn to alter
its own behaviour in response to rewards received. This paradigm of
trial-and error-learning has its roots in behaviourist psychology, and
is one of the main foundations of RL \citep{sutton1998reinforcement}.
The other key influence on RL is optimal control, which has lent the
mathematical formalisms (most notably dynamic programming
\citep{bellman1952theory}) that underpin the field.

In the RL set-up, an autonomous \emph{agent}, controlled by a machine
learning algorithm, observes a \emph{state} \(\mathbf{s}_t\) from its
\emph{environment} at timestep \(t\). The agent interacts with the
environment by taking an \emph{action} \(\mathbf{a}_t\) in state
\(\mathbf{s}_t\). When the agent takes an action, the environment and
the agent transition to a new state \(\mathbf{s}_{t+1}\) based on the
current state and the chosen action. The state is a sufficient statistic
of the environment and thereby comprises all the necessary information
for the agent to take the best action, which can include parts of the
agent, such as the position of its actuators and sensors. In the optimal
control literature, states and actions are often denoted by
\(\mathbf{x}_t\) and \(\mathbf{u}_t\), respectively.

The best sequence of actions is determined by the \emph{rewards}
provided by the environment. Every time the environment transitions to a
new state, it also provides a scalar reward \(r_{t+1}\) to the agent as
feedback. The goal of the agent is to learn a \emph{policy} (control
strategy) \(\pi\) that maximises the expected \emph{return} (cumulative,
discounted reward). Given a state, a policy returns an action to
perform; an \emph{optimal policy} is any policy that maximises the
expected return in the environment. In this respect, RL aims to solve
the same problem as optimal control. However, the challenge in RL is
that the agent needs to learn about the consequences of actions in the
environment by trial and error, as, unlike in optimal control, a model
of the state transition dynamics is not available to the agent. Every
interaction with the environment yields information, which the agent
uses to update its knowledge. This \emph{perception-action-learning
loop} is illustrated in Figure \ref{fig:pal}.

\subsection{Markov Decision Processes}\label{markov-decision-processes}

Formally, RL can be described as a Markov decision process (MDP), which
consists of:

\begin{itemize}
\tightlist
\item
  A set of states \(\mathcal{S}\), plus a distribution of starting
  states \(p(\mathbf{s}_0)\).
\item
  A set of actions \(\mathcal{A}\).
\item
  Transition dynamics
  \(\mathcal{T}(\mathbf{s}_{t+1}|\mathbf{s}_t, \mathbf{a}_t)\) that map
  a state-action pair at time \(t\) onto a distribution of states at
  time \(t+1\).
\item
  An immediate/instantaneous reward function
  \(\mathcal{R}(\mathbf{s}_t, \mathbf{a}_t, \mathbf{s}_{t+1})\).
\item
  A discount factor \(\gamma \in [0, 1]\), where lower values place more
  emphasis on immediate rewards.
\end{itemize}

In general, the policy \(\pi\) is a mapping from states to a probability
distribution over actions:
\(\pi:\mathcal{S} \rightarrow p(\mathcal{A} = \mathbf{a}|\mathcal{S})\).
If the MDP is \emph{episodic}, i.e., the state is reset after each
episode of length \(T\), then the sequence of states, actions and
rewards in an episode constitutes a \emph{trajectory} or \emph{rollout}
of the policy. Every rollout of a policy accumulates rewards from the
environment, resulting in the return
\(R = \sum_{t=0}^{T-1} \gamma^t r_{t+1}\). The goal of RL is to find an
optimal policy, \(\pi^*\), which achieves the maximum expected return
from all states:

\begin{equation}
\pi^* = \argmax_\pi\mathbb{E}[R|\pi].
\label{eq:expected return}
\end{equation}

It is also possible to consider non-episodic MDPs, where \(T = \infty\).
In this situation, \(\gamma < 1\) prevents an infinite sum of rewards
from being accumulated. Furthermore, methods that rely on complete
trajectories are no longer applicable, but those that use a finite set
of transitions still are.

A key concept underlying RL is the Markov property---only the current
state affects the next state, or in other words, the future is
conditionally independent of the past given the present state. This
means that any decisions made at \(\mathbf{s}_t\) can be based solely on
\(\mathbf{s}_{t-1}\), rather than
\(\{\mathbf{s}_0, \mathbf{s}_1, \ldots, \mathbf{s}_{t-1}\}\). Although
this assumption is held by the majority of RL algorithms, it is somewhat
unrealistic, as it requires the states to be \emph{fully observable}. A
generalisation of MDPs are partially observable MDPs (POMDPs), in which
the agent receives an observation \(\mathbf{o}_t \in \Omega\), where the
distribution of the observation
\(p(\mathbf{o}_{t+1}|\mathbf{s}_{t+1}, \mathbf{a}_t)\) is dependent on
the current state and the previous action \citep{kaelbling1998planning}.
In a control and signal processing context, the observation would be
described by a measurement\slash observation mapping in a
state-space-model that depends on the current state and the previously
applied action.

POMDP algorithms typically maintain a \emph{belief} over the current
state given the previous belief state, the action taken and the current
observation. A more common approach in deep learning is to utilise
recurrent neural networks (RNNs)
\citep{wierstra2010recurrent, hausknecht2015deep, heess2015memory, mnih2016asynchronous, oh2016control},
which, unlike feedforward neural networks, are dynamical systems. This
approach to solving POMDPs is related to other problems using dynamical
systems and state space models, where the true state can only be
estimated \citep{bertsekas2005dynamic}.

\subsection{Challenges in RL}\label{challenges-in-rl}

It is instructive to emphasise some challenges faced in RL:

\begin{itemize}
\tightlist
\item
  The optimal policy must be inferred by trial-and-error interaction
  with the environment. The only learning signal the agent receives is
  the reward.
\item
  The observations of the agent depend on its actions and can contain
  strong temporal correlations.
\item
  Agents must deal with long-range time dependencies: Often the
  consequences of an action only materialise after many transitions of
  the environment. This is known as the (temporal) \emph{credit
  assignment problem} \citep{sutton1998reinforcement}.
\end{itemize}

We will illustrate these challenges in the context of an indoor robotic
visual navigation task: if the goal location is specified, we may be
able to estimate the distance remaining (and use it as a reward signal),
but it is unlikely that we will know exactly what series of actions the
robot needs to take to reach the goal. As the robot must choose where to
go as it navigates the building, its decisions influence which rooms it
sees and, hence, the statistics of the visual sequence captured.
Finally, after navigating several junctions, the robot may find itself
in a dead end. There is a range of problems, from learning the
consequences of actions to balancing exploration against exploitation,
but ultimately these can all be addressed formally within the framework
of RL.

\section{Reinforcement Learning
Algorithms}\label{reinforcement-learning-algorithms}

So far, we have introduced the key formalism used in RL, the MDP, and
briefly noted some challenges in RL. In the following, we will
distinguish between different classes of RL algorithms. There are two
main approaches to solving RL problems: methods based on \emph{value
functions} and methods based on \emph{policy search}. There is also a
hybrid, \emph{actor-critic} approach, which employs both value functions
and policy search. We will now explain these approaches and other useful
concepts for solving RL problems.

\subsection{Value Functions}\label{value-functions}

Value function methods are based on estimating the value (expected
return) of being in a given state. The \emph{state-value function}
\(V^\pi(\mathbf{s})\) is the expected return when starting in state
\(\mathbf{s}\) and following \(\pi\) henceforth:

\begin{align}
V^\pi(\mathbf{s}) = \mathbb{E}[R|\mathbf{s}, \pi]
\label{eq:value + expected return}
\end{align}

The optimal policy, \(\pi^*\), has a corresponding state-value function
\(V^*(\mathbf{s})\), and vice-versa, the optimal state-value function
can be defined as

\begin{align}
V^*(\mathbf{s}) = \max_\pi V^\pi(\mathbf{s}) \quad \forall \mathbf{s} \in \mathcal{S}.
\label{eq:value function}
\end{align}

If we had \(V^*(\mathbf{s})\) available, the optimal policy could be
retrieved by choosing among all actions available at \(\mathbf{s}_t\)
and picking the action \(\mathbf{a}\) that maximises
\(\mathbb{E}_{\mathbf{s}_{t+1} \sim \mathcal{T}(\mathbf{s}_{t+1}|\mathbf{s}_t, \mathbf{a})}[V^*(\mathbf{s}_{t+1})]\).

In the RL setting, the transition dynamics \(\mathcal{T}\) are
unavailable. Therefore, we construct another function, the
\emph{state-action-value} or \emph{quality function}
\(Q^\pi(\mathbf{s}, \mathbf{a})\), which is similar to \(V^\pi\), except
that the initial action \(\mathbf{a}\) is provided, and \(\pi\) is only
followed from the succeeding state onwards:

\begin{align}
Q^\pi(\mathbf{s}, \mathbf{a}) = \mathbb{E}[R|\mathbf{s}, \mathbf{a}, \pi].
\label{eq:Q-function and expected return}
\end{align}

The best policy, given \(Q^\pi(\mathbf{s}, \mathbf{a})\), can be found
by choosing \(\mathbf{a}\) greedily at every state:
\(\argmax_\mathbf{a} Q^\pi(\mathbf{s}, \mathbf{a})\). Under this policy,
we can also define \(V^\pi(\mathbf{s})\) by maximising
\(Q^\pi(\mathbf{s}, \mathbf{a})\):
\(V^\pi(\mathbf{s}) = \max_\mathbf{a} Q^\pi(\mathbf{s}, \mathbf{a})\).

\textbf{Dynamic Programming:} To actually learn \(Q^\pi\), we exploit
the Markov property and define the function as a Bellman equation
\citep{bellman1952theory}, which has the following recursive form:

\begin{align}
Q^\pi(\mathbf{s}_t, \mathbf{a}_t) = \mathbb{E}_{\mathbf{s}_{t+1}}[r_{t+1} + \gamma Q^\pi(\mathbf{s}_{t+1}, \pi(\mathbf{s}_{t+1}))].
\end{align}

This means that \(Q^\pi\) can be improved by \emph{bootstrapping}, i.e.,
we can use the current values of our estimate of \(Q^\pi\) to improve
our estimate. This is the foundation of \(Q\)-learning
\citep{watkins1992q} and the state-action-reward-state-action (SARSA)
algorithm \citep{rummery1994line}:

\begin{align}
Q^\pi(\mathbf{s}_t, \mathbf{a}_t) \leftarrow Q^\pi(\mathbf{s}_t, \mathbf{a}_t) + \alpha \delta,
\end{align}

where \(\alpha\) is the learning rate and
\(\delta = Y - Q^\pi(\mathbf{s}_t, \mathbf{a}_t)\) the temporal
difference (TD) error; here, \(Y\) is a target as in a standard
regression problem. SARSA, an \emph{on-policy} learning algorithm, is
used to improve the estimate of \(Q^\pi\) by using transitions generated
by the behavioural policy (the policy derived from \(Q^\pi\)), which
results in setting
\(Y = r_t + \gamma Q^\pi(\mathbf{s}_{t+1}, \mathbf{a}_{t+1})\).
\(Q\)-learning is \emph{off-policy}, as \(Q^\pi\) is instead updated by
transitions that were not necessarily generated by the derived policy.
Instead, \(Q\)-learning uses
\(Y = r_t + \gamma\max_\mathbf{a} Q^\pi(\mathbf{s}_{t+1}, \mathbf{a})\),
which directly approximates \(Q^*\).

To find \(Q^*\) from an arbitrary \(Q^\pi\), we use \emph{generalised
policy iteration}, where policy iteration consists of \emph{policy
evaluation} and \emph{policy improvement}. Policy evaluation improves
the estimate of the value function, which can be achieved by minimising
TD errors from trajectories experienced by following the policy. As the
estimate improves, the policy can naturally be improved by choosing
actions greedily based on the updated value function. Instead of
performing these steps separately to convergence (as in policy
iteration), generalised policy iteration allows for interleaving the
steps, such that progress can be made more rapidly.

\subsection{Sampling}\label{sampling}

\begin{figure*}[ht]
  \begin{minipage}[t]{.49\textwidth}
    \centering
  \includegraphics[height=7cm]{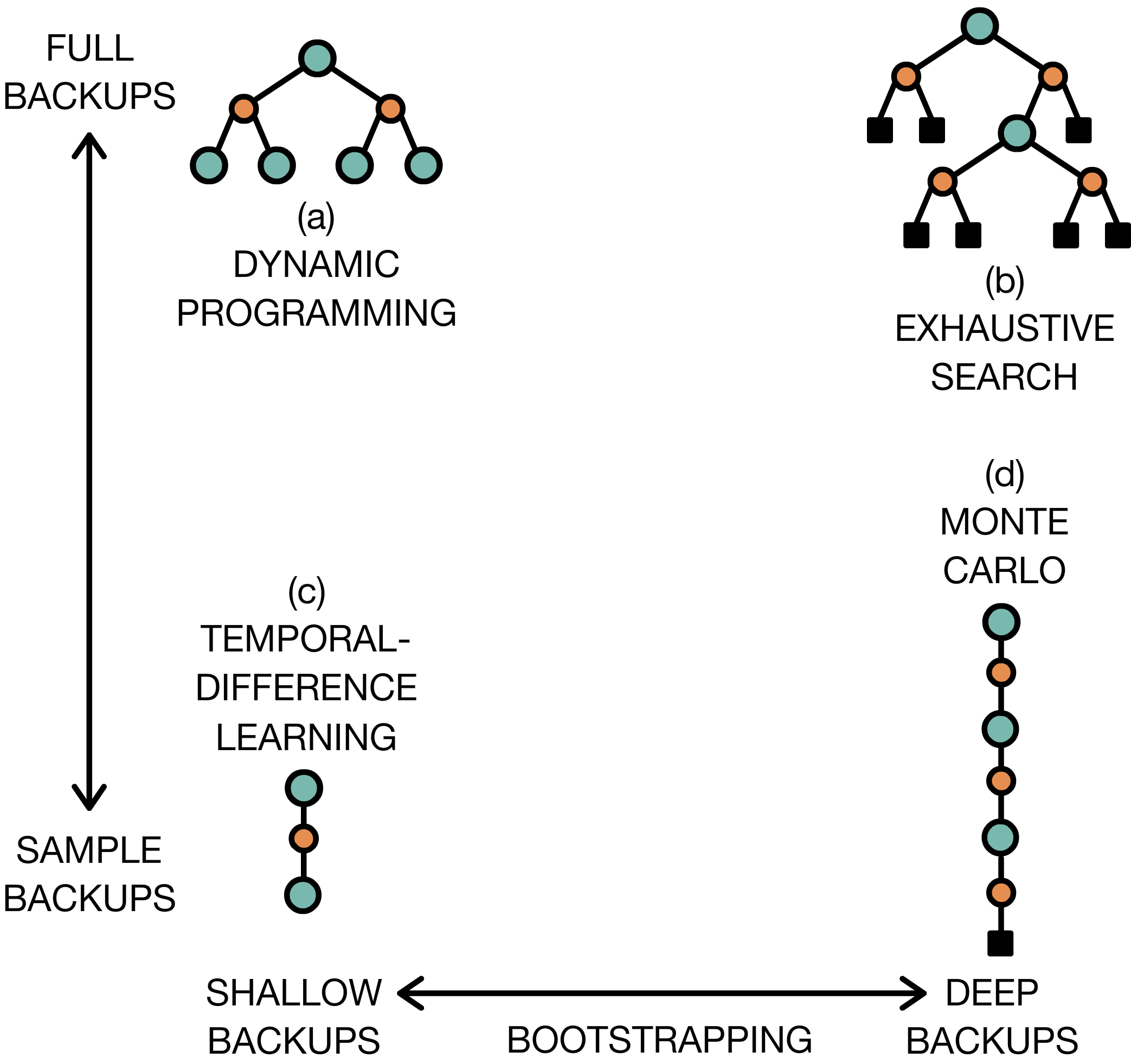}
  \caption{Two dimensions of RL algorithms, based on the \emph{backups} used to learn or construct a policy. At the extremes of these dimensions are (a) dynamic programming, (b) exhaustive search, (c) one-step TD learning and (d) pure Monte Carlo approaches. Bootstrapping extends from (c) 1-step TD learning to $n$-step TD learning methods \cite{sutton1998reinforcement}, with (d) pure Monte Carlo approaches not relying on bootstrapping at all. Another possible dimension of variation is choosing to (c, d) sample actions versus (a, b) taking the expectation over all choices. Recreated from \cite{sutton1998reinforcement}.}
  \label{fig:dimensions}
  \end{minipage}\hfill
  \begin{minipage}[t]{.49\textwidth}
    \centering
    \includegraphics[height=7cm]{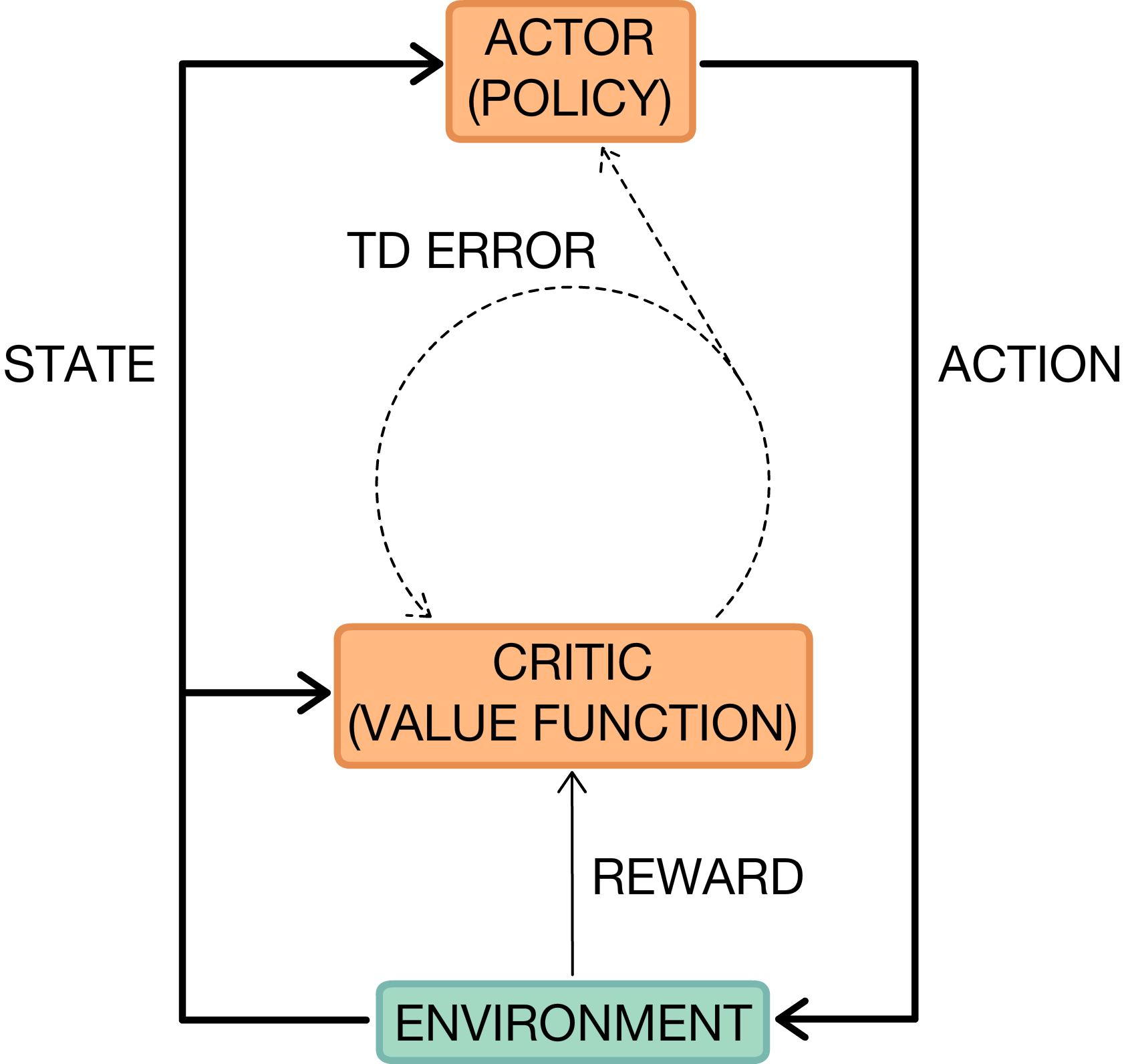}
    \caption{Actor-critic set-up. The actor (policy) receives a state from the environment and chooses an action to perform. At the same time, the critic (value function) receives the state and reward resulting from the previous interaction. The critic uses the TD error calculated from this information to update itself and the actor. Recreated from \cite{sutton1998reinforcement}.}
    \label{fig:actor-critic}
  \end{minipage}
\end{figure*}

Instead of bootstrapping value functions using dynamic programming
methods, Monte Carlo methods estimate the expected return
\eqref{eq:value + expected return} from a state by averaging the return
from multiple rollouts of a policy. Because of this, pure Monte Carlo
methods can also be applied in non-Markovian environments. On the other
hand, they can only be used in episodic MDPs, as a rollout has to
terminate for the return to be calculated. It is possible to get the
best of both methods by combining TD learning and Monte Carlo policy
evaluation, as in done in the TD(\(\lambda\)) algorithm
\citep{sutton1998reinforcement}. Similarly to the discount factor, the
\(\lambda\) in TD(\(\lambda\)) is used to interpolate between Monte
Carlo evaluation and bootstrapping. As demonstrated in Figure
\ref{fig:dimensions}, this results in an entire spectrum of RL methods
based around the amount of sampling utilised.

Another major value-function based method relies on learning the
\emph{advantage} function \(A^\pi(\mathbf{s}, \mathbf{a})\)
\citep{baird1993advantage, harmon1996multi}. Unlike producing absolute
state-action values, as with \(Q^\pi\), \(A^\pi\) instead represents
relative state-action values. Learning relative values is akin to
removing a baseline or average level of a signal; more intuitively, it
is easier to learn that one action has better consequences than another,
than it is to learn the actual return from taking the action. \(A^\pi\)
represents a relative advantage of actions through the simple
relationship \(A^\pi = Q^\pi - V^\pi\), and is also closely related to
the baseline method of variance reduction within gradient-based policy
search methods \citep{williams1992simple}. The idea of advantage updates
has been utilised in many recent DRL algorithms
\citep{wang2016dueling, gu2016continuous, mnih2016asynchronous, schulman2016high}.

\subsection{Policy Search}\label{policy-search}

Policy search methods do not need to maintain a value function model,
but directly search for an optimal policy \(\pi^*\). Typically, a
parameterised policy \(\pi_\theta\) is chosen, whose parameters are
updated to maximise the expected return \(\mathbb{E}[R|\theta]\) using
either gradient-based or gradient-free optimisation
\citep{deisenroth2013survey}. Neural networks that encode policies have
been successfully trained using both gradient-free
\citep{gomez2005evolving, cuccu2011intrinsically, koutnik2013evolving}
and gradient-based
\citep{williams1992simple, wierstra2010recurrent, heess2015learning, lillicrap2016continuous, schulman2015trust, schulman2016high, levine2016end}
methods. Gradient-free optimisation can effectively cover
low-dimensional parameter spaces, but despite some successes in applying
them to large networks \citep{koutnik2013evolving}, gradient-based
training remains the method of choice for most DRL algorithms, being
more sample-efficient when policies possess a large number of
parameters.

When constructing the policy directly, it is common to output parameters
for a probability distribution; for continuous actions, this could be
the mean and standard deviations of Gaussian distributions, whilst for
discrete actions this could be the individual probabilities of a
multinomial distribution. The result is a stochastic policy from which
we can directly sample actions. With gradient-free methods, finding
better policies requires a heuristic search across a predefined class of
models. Methods such as evolution strategies essentially perform
hill-climbing in a subspace of policies \citep{salimans2017evolution},
whilst more complex methods, such as compressed network search, impose
additional inductive biases \citep{koutnik2013evolving}. Perhaps the
greatest advantage of gradient-free policy search is that they can also
optimise non-differentiable policies.

\textbf{Policy Gradients:} Gradients can provide a strong learning
signal as to how to improve a parameterised policy. However, to compute
the expected return \eqref{eq:expected return} we need to average over
plausible trajectories induced by the current policy parameterisation.
This averaging requires either deterministic approximations (e.g.,
linearisation) or stochastic approximations via sampling
\citep{deisenroth2013survey}. Deterministic approximations can only be
applied in a model-based setting where a model of the underlying
transition dynamics is available. In the more common model-free RL
setting, a Monte Carlo estimate of the expected return is determined.
For gradient-based learning, this Monte Carlo approximation poses a
challenge since gradients cannot pass through these samples of a
stochastic function. Therefore, we turn to an estimator of the gradient,
known in RL as the REINFORCE rule \citep{williams1992simple}, elsewhere
known as the score function \citep{fu2006gradient} or likelihood-ratio
estimator \citep{glynn1990likelihood}. The latter name is telling as
using the estimator is similar to the practice of optimising the
log-likelihood in supervised learning. Intuitively, gradient ascent
using the estimator increases the log probability of the sampled action,
weighted by the return. More formally, the REINFORCE rule can be used to
compute the gradient of an expectation over a function \(f\) of a random
variable \(X\) with respect to parameters \(\theta\):

\begin{align}
\nabla_\theta\mathbb{E}_X[f(X; \theta)] = \mathbb{E}_X[f(X; \theta)\nabla_\theta\log p(X)].
\end{align}

As this computation relies on the empirical return of a trajectory, the
resulting gradients possess a high variance. By introducing unbiased
estimates that are less noisy it is possible to reduce the variance. The
general methodology for performing this is to subtract a baseline, which
means weighting updates by an advantage rather than the pure return. The
simplest baseline is the average return taken over several episodes
\citep{williams1992simple}, but many more options are available
\citep{schulman2016high}.

\textbf{Actor-critic Methods:} It is possible to combine value functions
with an explicit representation of the policy, resulting in actor-critic
methods, as shown in Figure \ref{fig:actor-critic}. The ``actor''
(policy) learns by using feedback from the ``critic'' (value function).
In doing so, these methods trade off variance reduction of policy
gradients with bias introduction from value function methods
\citep{konda2003onactor, schulman2016high}.

Actor-critic methods use the value function as a baseline for policy
gradients, such that the only fundamental difference between
actor-critic methods and other baseline methods are that actor-critic
methods utilise a \emph{learned} value function. For this reason, we
will later discuss actor-critic methods as a subset of policy gradient
methods.

\subsection{Planning and Learning}\label{planning-and-learning}

Given a model of the environment, it is possible to use dynamic
programming over all possible actions (Figure \ref{fig:dimensions} (a)),
sample trajectories for heuristic search (as was done by AlphaGo
\citep{silver2016mastering}), or even perform an exhaustive search
(Figure \ref{fig:dimensions} (b)). Sutton and Barto
\citep{sutton1998reinforcement} define \emph{planning} as any method
which utilises a model to produce or improve a policy. This includes
\emph{distribution models}, which include \(\mathcal{T}\) and
\(\mathcal{R}\), and \emph{sample models}, from which only samples of
transitions can be drawn.

In RL, we focus on learning without access to the underlying model of
the environment. However, interactions with the environment could be
used to learn value functions, policies, and also a model. Model-free RL
methods learn directly from interactions with the environment, but
model-based RL methods can simulate transitions using the learned model,
resulting in increased sample efficiency. This is particularly important
in domains where each interaction with the environment is expensive.
However, learning a model introduces extra complexities, and there is
always the danger of suffering from model errors, which in turn affects
the learned policy; a common but partial solution in this latter
scenario is to use model predictive control, where planning is repeated
after small sequences of actions in the real environment
\citep{bertsekas2005dynamic}. Although deep neural networks can
potentially produce very complex and rich models
\citep{oh2015action, stadie2015incentivizing, finn2016deep}, sometimes
simpler, more data-efficient methods are preferable
\citep{gu2016continuous}. These considerations also play a role in
actor-critic methods with learned value functions
\citep{konda2003onactor, schulman2016high}.

\subsection{The Rise of DRL}\label{the-rise-of-drl}

Many of the successes in DRL have been based on scaling up prior work in
RL to high-dimensional problems. This is due to the learning of
low-dimensional feature representations and the powerful function
approximation properties of neural networks. By means of representation
learning, DRL can deal efficiently with the curse of dimensionality,
unlike tabular and traditional non-parametric methods
\citep{bengio2013representation}. For instance, convolutional neural
networks (CNNs) can be used as components of RL agents, allowing them to
learn directly from raw, high-dimensional visual inputs. In general, DRL
is based on training deep neural networks to approximate the optimal
policy \(\pi^*\), and/or the optimal value functions \(V^*\), \(Q^*\)
and \(A^*\).

Although there have been DRL successes with gradient-free methods
\citep{gomez2005evolving, cuccu2011intrinsically, koutnik2013evolving},
the vast majority of current works rely on gradients and hence the
backpropagation algorithm
\citep{werbos1974beyond, rumelhart1988learning}. The primary motivation
is that when available, gradients provide a strong learning signal. In
reality, these gradients are estimated based on approximations, through
sampling or otherwise, and as such we have to craft algorithms with
useful inductive biases in order for them to be tractable.

The other benefit of backpropagation is to view the optimisation of the
expected return as the optimisation of a stochastic function
\citep{schulman2015gradient, heess2015learning}. This function can
comprise of several parts---models, policies and value functions---which
can be combined in various ways. The individual parts, such as value
functions, may not directly optimise the expected return, but can
instead embody useful information about the RL domain. For example,
using a differentiable model and policy, it is possible to forward
propagate and backpropagate through entire rollouts; on the other hand,
innacuracies can accumulate over long time steps, and it may be be
pertinent to instead use a value function to summarise the statistics of
the rollouts \citep{heess2015learning}. We have previously mentioned
that representation learning and function approximation are key to the
success of DRL, but it is also true to say that the field of deep
learning has inspired new ways of thinking about RL.

Following our review of RL, we will now partition the next part of the
survey into value function and policy search methods in DRL, starting
with the well-known deep \(Q\)-network (DQN) \citep{mnih2015human}. In
these sections, we will focus on state-of-the-art techniques, as well as
the historical works they are built upon. The focus of the
state-of-the-art techniques will be on those for which the state space
is conveyed through visual inputs, e.g., images and video. To conclude,
we will examine ongoing research areas and open challenges.

\section{Value Functions}\label{value-functions-1}

The well-known function approximation properties of neural networks led
naturally to the use of deep learning to regress functions for use in RL
agents. Indeed, one of the earliest success stories in RL is TD-Gammon,
a neural network that reached expert-level performance in Backgammon in
the early 90s \citep{tesauro1995temporal}. Using TD methods, the network
took in the state of the board to predict the probability of black or
white winning. Although this simple idea has been echoed in later work
\citep{silver2016mastering}, progress in RL research has favoured the
explicit use of value functions, which can capture the structure
underlying the environment. From early value function methods in DRL,
which took simple states as input \citep{riedmiller2005neural}, current
methods are now able to tackle visually and conceptually complex
environments
\citep{mnih2015human, schulman2015trust, mnih2016asynchronous, oh2016control, zhu2017target}.

\subsection{Function Approximation and the
DQN}\label{function-approximation-and-the-dqn}

\begin{figure*}[ht]
  \centering
  \includegraphics[width=0.75\textwidth]{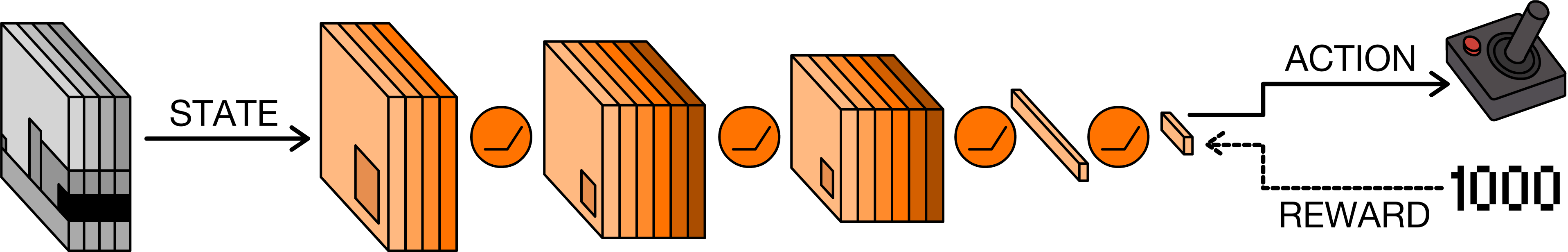}
  \caption{The deep $Q$-network \cite{mnih2015human}. The network takes the state---a stack of greyscale frames from the video game---and processes it with convolutional and fully connected layers, with ReLU nonlinearities in between each layer. At the final layer, the network outputs a discrete action, which corresponds to one of the possible control inputs for the game. Given the current state and chosen action, the game returns a new score. The DQN uses the reward---the difference between the new score and the previous one---to learn from its decision. More precisely, the reward is used to update its estimate of $Q$, and the error between its previous estimate and its new estimate is backpropagated through the network.}
  \label{fig:dqn}
\end{figure*}

We begin our survey of value-function-based DRL algorithms with the DQN
\citep{mnih2015human}, pictured in Figure \ref{fig:dqn}, which achieved
scores across a wide range of classic Atari 2600 video games
\citep{bellemare2015arcade} that were comparable to that of a
professional video games tester. The inputs to the DQN are four
greyscale frames of the game, concatenated over time, which are
initially processed by several convolutional layers in order to extract
spatiotemporal features, such as the movement of the ball in ``Pong'' or
``Breakout.'' The final feature map from the convolutional layers is
processed by several fully connected layers, which more implicitly
encode the effects of actions. This contrasts with more traditional
controllers that use fixed preprocessing steps, which are therefore
unable to adapt their processing of the state in response to the
learning signal.

A forerunner of the DQN---neural fitted \(Q\) iteration (NFQ)---involved
training a neural network to return the \(Q\)-value given a state-action
pair \citep{riedmiller2005neural}. NFQ was later extended to train a
network to drive a slot car using raw visual inputs from a camera over
the race track, by combining a deep autoencoder to reduce the
dimensionality of the inputs with a separate branch to predict
\(Q\)-values \citep{lange2012autonomous}. Although the previous network
could have been trained for both reconstruction and RL tasks
simultaneously, it was both more reliable and computationally efficient
to train the two parts of the network sequentially.

\begin{figure}[ht]
  \centering
  \includegraphics[height=5.5cm]{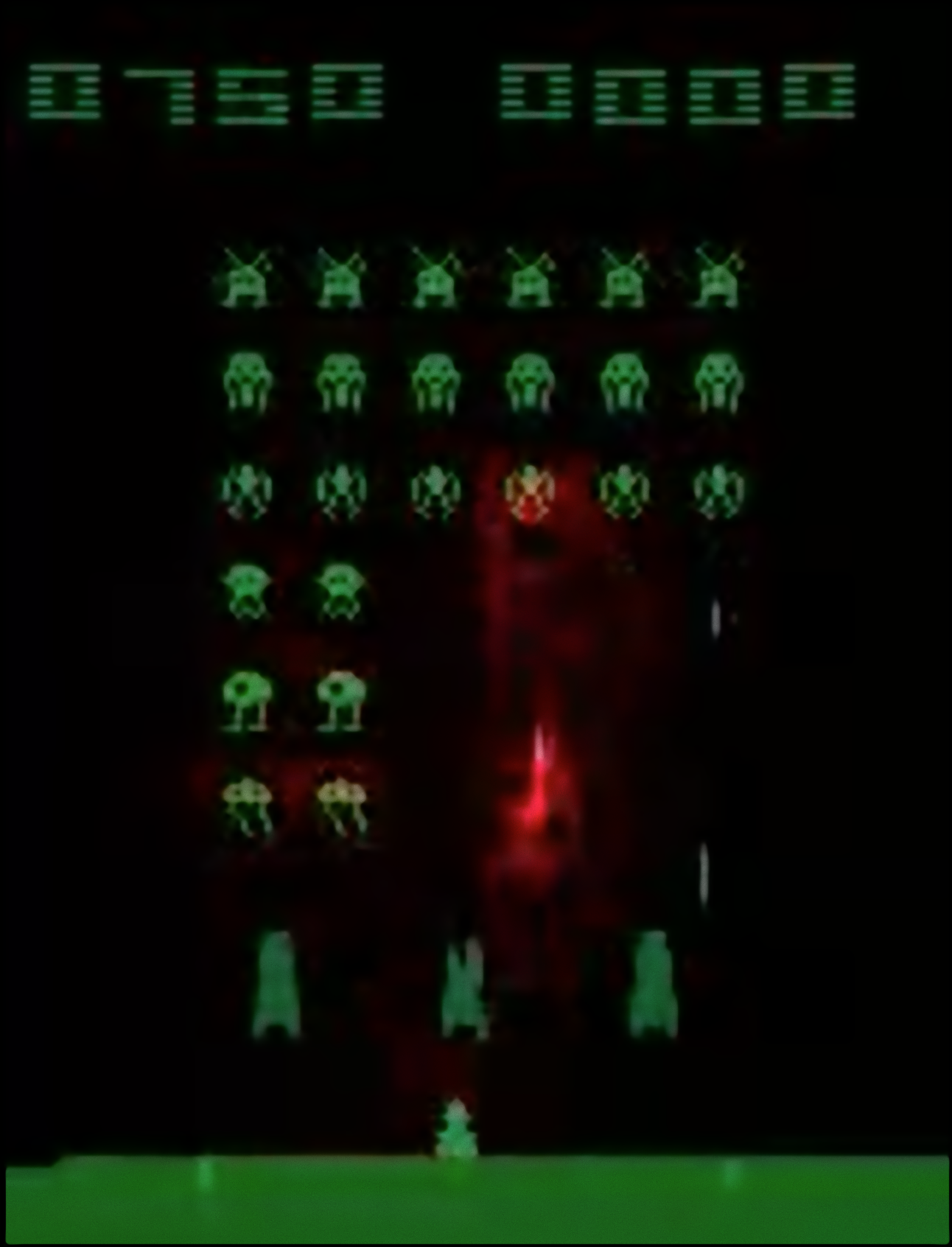}
  \caption{Saliency map of a trained DQN \cite{mnih2015human} playing ``Space Invaders'' \cite{bellemare2015arcade}. By backpropagating the training signal to the image space, it is possible to see what a neural-network-based agent is attending to. In this frame, the most salient points---shown with the red overlay---are the laser that the agent recently fired, and also the enemy that it anticipates hitting in a few time steps.}
  \label{fig:saliency}
\end{figure}

The DQN \citep{mnih2015human} is closely related to the model proposed
by Lange et al. \citep{lange2012autonomous}, but was the first RL
algorithm that was demonstrated to work directly from raw visual inputs
and on a wide variety of environments. It was designed such that the
final fully connected layer outputs \(Q^\pi(\mathbf{s}, \cdot)\) for all
action values in a discrete set of actions---in this case, the various
directions of the joystick and the fire button. This not only enables
the best action, \(\argmax_\mathbf{a} Q^\pi(\mathbf{s}, \mathbf{a})\),
to be chosen after a single forward pass of the network, but also allows
the network to more easily encode action-independent knowledge in the
lower, convolutional layers. With merely the goal of maximising its
score on a video game, the DQN learns to extract salient visual
features, jointly encoding objects, their movements, and, most
importantly, their interactions. Using techniques originally developed
for explaining the behaviour of CNNs in object recognition tasks, we can
also inspect what parts of its view the agent considers important (see
Figure \ref{fig:saliency}).

The true underlying state of the game is contained within 128 bytes of
Atari 2600 RAM. However, the DQN was designed to directly learn from
visual inputs (\(210 \times 160 \text{pixel}\) 8-bit RGB images), which
it takes as the state \(\mathbf{s}\). It is impractical to represent
\(Q^\pi(\mathbf{s}, \mathbf{a})\) exactly as a lookup table: When
combined with 18 possible actions, we obtain a \(Q\)-table of size
\(|\mathcal{S}| \times |\mathcal{A}| = 18 \times 256^{3 \times 210 \times 160}\).
Even if it were feasible to create such a table, it would be sparsely
populated, and information gained from one state-action pair cannot be
propagated to other state-action pairs. The strength of the DQN lies in
its ability to compactly represent both high-dimensional observations
and the \(Q\)-function using deep neural networks. Without this ability,
tackling the discrete Atari domain from raw visual inputs would be
impractical.

The DQN addressed the fundamental instability problem of using function
approximation in RL \citep{tsitsiklis1997analysis} by the use of two
techniques: experience replay \citep{lin1992self} and target networks.
Experience replay memory stores transitions of the form
\((\mathbf{s}_t, \mathbf{a}_t, \mathbf{s}_{t+1}, r_{t+1})\) in a cyclic
buffer, enabling the RL agent to sample from and train on previously
observed data offline. Not only does this massively reduce the amount of
interactions needed with the environment, but batches of experience can
be sampled, reducing the variance of learning updates. Furthermore, by
sampling uniformly from a large memory, the temporal correlations that
can adversely affect RL algorithms are broken. Finally, from a practical
perspective, batches of data can be efficiently processed in parallel by
modern hardware, increasing throughput. Whilst the original DQN
algorithm used uniform sampling \citep{mnih2015human}, later work showed
that prioritising samples based on TD errors is more effective for
learning \citep{schaul2016prioritized}. We note that although experience
replay is typically thought of as a model-free technique, it could
actually be considered a simple model \citep{vanseijen2015deeper}.

The second stabilising method, introduced by Mnih et al.
\citep{mnih2015human}, is the use of a target network that initially
contains the weights of the network enacting the policy, but is kept
frozen for a large period of time. Rather than having to calculate the
TD error based on its own rapidly fluctuating estimates of the
\(Q\)-values, the policy network uses the fixed target network. During
training, the weights of the target network are updated to match the
policy network after a fixed number of steps. Both experience replay and
target networks have gone on to be used in subsequent DRL works
\citep{gu2016continuous, lillicrap2016continuous, wang2017sample, nachum2017bridging}.

\subsection{\texorpdfstring{\(Q\)-Function
Modifications}{Q-Function Modifications}}\label{q-function-modifications}

Considering that one of the key components of the DQN is a function
approximator for the \(Q\)-function, it can benefit from fundamental
advances in RL. van Hasselt \citep{hasselt2010double} showed that the
single estimator used in the \(Q\)-learning update rule overestimates
the expected return due to the use of the maximum action value as an
approximation of the maximum \emph{expected} action value. Double-\(Q\)
learning provides a better estimate through the use of a double
estimator \citep{hasselt2010double}. Whilst double-\(Q\) learning
requires an additional function to be learned, later work proposed using
the already available target network from the DQN algorithm, resulting
in significantly better results with only a small change in the update
step \citep{van2016deep}. A more radical proposal by Bellemare et al.
\citep{bellemare2017distributional} was to actually learn the full
\emph{value distribution}, rather than just the expectation; this
provides additional information, such as whether the potential rewards
come from a skewed or multimodal distribution. Although the resulting
algorithm---based on learning categorical distributions---was used to
construct the Categorical DQN, the benefits can potentially be applied
to any RL algorithm that utilises learned value functions.

Yet another way to adjust the DQN architecture is to decompose the
\(Q\)-function into meaningful functions, such as constructing \(Q^\pi\)
by adding together separate layers that compute the state-value function
\(V^\pi\) and advantage function \(A^\pi\) \citep{wang2016dueling}.
Rather than having to come up with accurate \(Q\)-values for all
actions, the duelling DQN \citep{wang2016dueling} benefits from a single
baseline for the state in the form of \(V^\pi\), and easier-to-learn
relative values in the form of \(A^\pi\). The combination of the
duelling DQN with prioritised experience replay
\citep{schaul2016prioritized} is one of the state-of-the-art techniques
in discrete action settings. Further insight into the properties of
\(A^\pi\) by Gu et al. \citep{gu2016continuous} led them to modify the
DQN with a convex advantage layer that extended the algorithm to work
over sets of continuous actions, creating the normalised advantage
function (NAF) algorithm. Benefiting from experience replay, target
networks and advantage updates, NAF is one of several state-of-the-art
techniques in continuous control problems \citep{gu2016continuous}.

Some RL domains, such as recommender systems, have very large discrete
action spaces, and hence may be difficult to directly deal with.
Dulac-Arnold et al. \citep{dulac2015deep} proposed learning ``action
embeddings'' over the large set of original actions, and then using
\(k\)-nearest neighbors to produce ``proto-actions'' which can be used
with traditional RL methods. The idea of using representation learning
to create distributed embeddings is a particular strength of DRL, and
has been successfully utilised for other purposes
\citep{weber2017imagination, pascanu2017learning}. Another related
scenario in RL is when many actions need to be made simultaneously, such
as specifying the torques in a many-jointed robot, which results in the
action space growing exponentially. A naive but reasonable approach is
to factorise the policy, treating each action independently
\citep{rusu2017sim}. An alternative is to construct an autoregressive
policy, where each action in a single timestep is predicted
conditionally on the state and previously chosen actions from the same
timestep \citep{ranzato2016sequence, bahdanau2017actor, zoph2017neural}.
Metz et al. \citep{metz2017discrete} used this idea in order to
construct the sequential DQN, allowing them to discretise a large action
space and outperform NAF---which is limited by its quadratic advantage
function---in continous control problems. In a broader context, rather
than dealing directly with primitive actions directly, one may choose to
invoke ``subpolicies'' from higher-level policies
\citep{sutton1999between}; this concept, known as hierarchical
reinforcement learning (HRL), will be discussed later.

\section{Policy Search}\label{policy-search-1}

Policy search methods aim to directly find policies by means of
gradient-free or gradient-based methods. Prior to the current surge of
interest in DRL, several successful methods in DRL eschewed the commonly
used backpropagation algorithm in favour of evolutionary algorithms
\citep{gomez2005evolving, cuccu2011intrinsically, koutnik2013evolving},
which are gradient-free policy search algorithms. Evolutionary methods
rely on evaluating the performance of a population of agents. Hence,
they are expensive for large populations or agents with many parameters.
However, as black-box optimisation methods they can be used to optimise
arbitrary, non-differentiable models and naturally allow for more
exploration in parameter space. In combination with a compressed
representation of neural network weights, evolutionary algorithms can
even be used to train large networks; such a technique resulted in the
first deep neural network to learn an RL task, straight from
high-dimensional visual inputs \citep{koutnik2013evolving}. Recent work
has reignited interest in evolutionary methods for RL as they can
potentially be distributed at larger scales than techniques that rely on
gradients \citep{salimans2017evolution}.

\subsection{Backpropagation through Stochastic
Functions}\label{backpropagation-through-stochastic-functions}

The workhorse of DRL, however, remains backpropagation
\citep{werbos1974beyond, rumelhart1988learning}. The previously
discussed REINFORCE rule \citep{williams1992simple} allows neural
networks to learn stochastic policies in a task-dependent manner, such
as deciding where to look in an image to track
\citep{schmidhuber1991learning}, classify \citep{mnih2014recurrent} or
caption objects \citep{xu2015show}. In these cases, the stochastic
variable would determine the coordinates of a small crop of the image,
and hence reduce the amount of computation needed. This usage of RL to
make discrete, stochastic decisions over inputs is known in the deep
learning literature as \emph{hard attention}, and is one of the more
compelling uses of basic policy search methods in recent years, having
many applications outside of traditional RL domains. More generally, the
ability to backpropagate through stochastic functions, using techniques
such as REINFORCE \citep{williams1992simple} or the ``reparameterisation
trick'' \citep{kingma2014auto, rezende2014stochastic}, allows neural
networks to be treated as stochastic computation graphs that can be
optimised over \citep{schulman2015gradient}, which is a key concept in
algorithms such as stochastic value gradients (SVGs)
\citep{heess2015learning}.

\subsection{Compounding Errors}\label{compounding-errors}

Searching directly for a policy represented by a neural network with
very many parameters can be difficult and can suffer from severe local
minima. One way around this is to use guided policy search (GPS), which
takes a few sequences of actions from another controller (which could be
constructed using a separate method, such as optimal control). GPS
learns from them by using supervised learning in combination with
importance sampling, which corrects for off-policy samples
\citep{levine2013guided}. This approach effectively biases the search
towards a good (local) optimum. GPS works in a loop, by optimising
policies to match sampled trajectories, and optimising trajectory
distributions to match the policy and minimise costs. Initially, GPS was
used to train neural networks on simulated continuous RL problems
\citep{levine2014learning}, but was later utilised to train a policy for
a real robot based on visual inputs \citep{levine2016end}. This research
by Levine et al. \citep{levine2016end} showed that it was possible to
train visuomotor policies for a robot ``end-to-end'', straight from the
RGB pixels of the camera to motor torques, and, hence, is one of the
seminal works in DRL.

A more commonly used method is to use a trust region, in which
optimisation steps are restricted to lie within a region where the
approximation of the true cost function still holds. By preventing
updated policies from deviating too wildly from previous policies, the
chance of a catastrophically bad update is lessened, and many algorithms
that use trust regions guarantee or practically result in monotonic
improvement in policy performance. The idea of constraining each policy
gradient update, as measured by the Kullback-Leibler (KL) divergence
between the current and proposed policy, has a long history in RL
\citep{kakade2002natural, bagnell2003covariant, kappen2005path, peters2010relative}.
One of the newer algorithms in this line of work, trust region policy
optimisation (TRPO), has been shown to be relatively robust and
applicable to domains with high-dimensional inputs
\citep{schulman2015trust}. To achieve this, TRPO optimises a
\emph{surrogate} objective function---specifically, it optimises an
(importance sampled) advantage estimate, constrained using a quadratic
approximation of the KL divergence. Whilst TRPO can be used as a pure
policy gradient method with a simple baseline, later work by Schulman et
al. \citep{schulman2016high} introduced generalised advantage estimation
(GAE), which proposed several, more advanced variance reduction
baselines. The combination of TRPO and GAE remains one of the
state-of-the-art RL techniques in continuous control. However, the
constrained optimisation of TRPO requires calculating second-order
gradients, limiting its applicability. In contrast, the newer proximal
policy optimisation (PPO) algorithm performs unconstrained optimisation,
requiring only first-order gradient information
\citep{abbeel2016deep, heess2017emergence, schulman2017proximal}. The
two main variants include an adaptive penalty on the KL divergence, and
a heuristic clipped objective which is independent of the KL divergence
\citep{schulman2017proximal}. Being less expensive whilst retaining the
performance of TRPO means that PPO (with or without GAE) is gaining
popularity for a range of RL tasks
\citep{heess2017emergence, schulman2017proximal}.

\subsection{Actor-Critic Methods}\label{actor-critic-methods}

Instead of utilising the average of several Monte Carlo returns as the
baseline for policy gradient methods, actor-critic approaches have grown
in popularity as an effective means of combining the benefits of policy
search methods with learned value functions, which are able to learn
from full returns and/or TD errors. They can benefit from improvements
in both policy gradient methods, such as GAE \citep{schulman2016high},
and value function methods, such as target networks
\citep{mnih2015human}. In the last few years, DRL actor-critic methods
have been scaled up from learning simulated physics tasks
\citep{heess2015learning, lillicrap2016continuous} to real robotic
visual navigation tasks \citep{zhu2017target}, directly from image
pixels.

One recent development in the context of actor-critic algorithms are
deterministic policy gradients (DPGs) \citep{silver2014deterministic},
which extend the standard policy gradient theorems for stochastic
policies \citep{williams1992simple} to deterministic policies. One of
the major advantages of DPGs is that, whilst stochastic policy gradients
integrate over both state and action spaces, DPGs only integrate over
the state space, requiring fewer samples in problems with large action
spaces. In the initial work on DPGs, Silver et al.
\citep{silver2014deterministic} introduced and demonstrated an
off-policy actor-critic algorithm that vastly improved upon a stochastic
policy gradient equivalent in high-dimensional continuous control
problems. Later work introduced deep DPG (DDPG), which utilised neural
networks to operate on high-dimensional, visual state spaces
\citep{lillicrap2016continuous}. In the same vein as DPGs, Heess et al.
\citep{heess2015learning} devised a method for calculating gradients to
optimise stochastic policies, by ``reparameterising''
\citep{kingma2014auto, rezende2014stochastic} the stochasticity away
from the network, thereby allowing standard gradients to be used
(instead of the high-variance REINFORCE estimator
\citep{williams1992simple}). The resulting SVG methods are flexible, and
can be used both with (SVG(0) and SVG(1)) and without (SVG(\(\infty\)))
value function critics, and with (SVG(\(\infty\)) and SVG(1)) and
without (SVG(0)) models. Later work proceeded to integrate DPGs and SVGs
with RNNs, allowing them to solve continuous control problems in POMDPs,
learning directly from pixels \citep{heess2015memory}.

Value functions introduce a broadly applicable benefit in actor-critic
methods---the ability to use off-policy data. On-policy methods can be
more stable, whilst off-policy methods can be more data efficient, and
hence there have been several attempts to merge the two
\citep{wang2017sample, o2017pgq, gu2017q, gruslys2017reactor, gu2017interpolated}.
Earlier work has either utilised a mix of on-policy and off-policy
gradient updates \citep{wang2017sample, o2017pgq, gruslys2017reactor},
or used the off-policy data to train a value function in order to reduce
the variance of on-policy gradient updates \citep{gu2017q}. The more
recent work by Gu et al. \citep{gu2017interpolated} unified these
methods under interpolated policy gradients (IPGs), resulting in one of
the newest state-of-the-art continuous DRL algorithms, and also
providing insights for future research in this area. Together, the ideas
behind IPGs and SVGs (of which DPGs can be considered a special case)
form algorithmic approaches for improving learning efficiency in DRL.

An orthogonal approach to speeding up learning is to exploit parallel
computation. In particular, methods for training networks through
asynchronous gradient updates have been developed for use on both single
machines \citep{recht2011hogwild} and distributed systems
\citep{dean2012large}. By keeping a canonical set of parameters that are
read by and updated in an asynchronous fashion by multiple copies of a
single network, computation can be efficiently distributed over both
processing cores in a single CPU, and across CPUs in a cluster of
machines. Using a distributed system, Nair et al.
\citep{nair2015massively} developed a framework for training multiple
DQNs in parallel, achieving both better performance and a reduction in
training time. However, the simpler asynchronous advantage actor-critic
(A3C) algorithm \citep{mnih2016asynchronous}, developed for both single
and distributed machine settings, has become one of the most popular DRL
techniques in recent times. A3C combines advantage updates with the
actor-critic formulation, and relies on asynchronously updated policy
and value function networks trained in parallel over several processing
threads. The use of multiple agents, situated in their own, independent
environments, not only stabilises improvements in the parameters, but
conveys an additional benefit in allowing for more exploration to occur.
A3C has been used as a standard starting point in many subsequent works,
including the work of Zhu et al. \citep{zhu2017target}, who applied it
to robotic navigation in the real world through visual inputs. For
simplicity, the underlying algorithm may be used with just one agent,
termed advantage actor-critic (A2C) \citep{wang2017learning}.
Alternatively, segments from the trajectories of multiple agents can be
collected and processed together in a batch, with batch processing more
efficiently enabled by GPUs; this synchronous version also goes by the
name of A2C \citep{schulman2017proximal}.

There have been several major advancements on the original A3C algorithm
that reflect various motivations in the field of DRL. The first is
actor-critic with experience replay
\citep{wang2017sample, gruslys2017reactor}, which adds
Retrace(\(\lambda\)) off-policy bias correction \citep{munos2016safe} to
a \(Q\)-value-based A3C, allowing it to use experience replay in order
to improve sample complexity. Others have attempted to bridge the gap
between value and policy-based RL, utilising theoretical advancements to
improve upon the original A3C
\citep{nachum2017bridging, o2017pgq, schulman2017equivalence}. Finally,
there is a growing trend towards exploiting auxiliary tasks to improve
the representations learned by DRL agents, and, hence, improve both the
learning speed and final performance of these agents
\citep{li2015recurrent, jaderberg2017reinforcement, mirowski2017learning}.

\section{Current Research and
Challenges}\label{current-research-and-challenges}

To conclude, we will highlight some current areas of research in DRL,
and the challenges that still remain. Previously, we have focused mainly
on model-free methods, but we will now examine a few model-based DRL
algorithms in more detail. Model-based RL algorithms play an important
role in making RL data-efficient and in trading off exploration and
exploitation. After tackling exploration strategies, we shall then
address HRL, which imposes an inductive bias on the final policy by
explicitly factorising it into several levels. When available,
trajectories from other controllers can be used to bootstrap the
learning process, leading us to imitation learning and inverse RL (IRL).
For the final topic specific to RL, we will look at multi-agent systems,
which have their own special considerations. We then bring to attention
two broader areas---the use of RNNs, and transfer learning---in the
context of DRL. We then examine the issue of evaluating RL, and current
benchmarks for DRL.

\subsection{Model-based RL}\label{model-based-rl}

The key idea behind model-based RL is to learn a transition model that
allows for simulation of the environment without interacting with the
environment directly. Model-based RL does not assume specific prior
knowledge. However, in practice, we can incorporate prior knowledge
(e.g., physics-based models \citep{kansky2017schema}) to speed up
learning. Model learning plays an important role in reducing the amount
of required interactions with the (real) environment, which may be
limited in practice. For example, it is unrealistic to perform millions
of experiments with a robot in a reasonable amount of time and without
significant hardware wear and tear. There are various approaches to
learn predictive models of dynamical systems using pixel information.
Based on the deep dynamical model \citep{wahlstrom2015learning}, where
high-dimensional observations are embedded into a lower-dimensional
space using autoencoders, several model-based DRL algorithms have been
proposed for learning models and policies from pixel information
\citep{oh2015action, watter2015embed, wahlstrom2015pixels}. If a
sufficiently accurate model of the environment can be learned, then even
simple controllers can be used to control a robot directly from camera
images \citep{finn2016deep}. Learned models can also be used to guide
exploration purely based on simulation of the environment, with deep
models allowing these techniques to be scaled up to high-dimensional
visual domains \citep{stadie2015incentivizing}.

A compelling insight on the benefits of neural-network-based models is
that they can overcome some of the problems incurred by planning with
imperfect models; in effect, by \emph{embedding} the activations and
predictions (outputs) of these models into a vector, a DRL agent can not
only obtain more information than just the final result of any model
rollouts, but it can also learn to downplay this information if it
believes that the model is inaccurate \citep{weber2017imagination}. This
can be more efficient, though less principled, than Bayesian methods for
propagating uncertainty \citep{houthooft2016vime}. Another way to make
use of the flexiblity of neural-network-based models is to let them
decide when to plan, that is, given a finite amount of computation,
whether it is worth modelling one long trajectory, several short
trajectories, anything in-between, or simply to take an action in the
real environment \citep{pascanu2017learning}.

Although deep neural networks can make reasonable predictions in
simulated environments over hundreds of timesteps
\citep{chiappa2017recurrent}, they typically require many samples to
tune the large amount of parameters they contain. Training these models
often requires more samples (interaction with the environment) than
simpler models. For this reason, Gu et al. \citep{gu2016continuous}
train locally linear models for use with the NAF algorithm---the
continuous equivalent of the DQN \citep{mnih2015human}---to improve the
algorithm's sample complexity in the robotic domain where samples are
expensive. In order to spur the adoption of deep models in model-based
DRL, it is necessary to find strategies that can be used in order to
improve their data efficiency \citep{nagabandi2017neural}.

A less common but potentially useful paradigm exists between model-free
and model-based methods---the successor representation (SR)
\citep{dayan1993improving}. Rather than picking actions directly or
performing planning with models, learning \(\mathcal{T}\) is replaced
with learning expected (discounted) future occupancies (SRs), which can
be linearly combined with \(\mathcal{R}\) in order to calculate the
optimal action; this decomposition makes SRs more robust than model-free
methods when the reward structure changes (but still fallible when
\(\mathcal{T}\) changes). Work extending SRs to deep neural networks has
demonstrated its usefulness in multi-task settings, whilst within a
complex visual environment \citep{kulkarni2016deep}.

\subsection{Exploration
vs.~Exploitation}\label{exploration-vs.exploitation}

One of the greatest difficulties in RL is the fundamental dilemma of
\emph{exploration versus exploitation}: When should the agent try out
(perceived) non-optimal actions in order to explore the environment (and
potentially improve the model), and when should it exploit the optimal
action in order to make useful progress? Off-policy algorithms, such as
the DQN \citep{mnih2015human}, typically use the simple
\(\epsilon\)-greedy exploration policy, which chooses a random action
with probability \(\epsilon\in [0,1]\), and the optimal action
otherwise. By decreasing \(\epsilon\) over time, the agent progresses
towards exploitation. Although adding independent noise for exploration
is usable in continuous control problems, more sophisticated strategies
inject noise that is correlated over time (e.g., from stochastic
processes) in order to better preserve momentum
\citep{lillicrap2016continuous}.

The observation that temporal correlation is important led Osband et al.
\citep{osband2016deep} to propose the bootstrapped DQN, which maintains
several \(Q\)-value ``heads'' that learn different values through a
combination of different weight initialisations and bootstrapped
sampling from experience replay memory. At the beginning of each
training episode, a different head is chosen, leading to
temporally-extended exploration. Usunier et al.
\citep{usunier2017episodic} later proposed a similar method that
performed exploration in policy space by adding noise to a single output
head, using zero-order gradient estimates to allow backpropagation
through the policy.

One of the main principled exploration strategies is the \emph{upper
confidence bound} (UCB) algorithm, based on the principle of ``optimism
in the face of uncertainty'' \citep{lai1985asymptotically}. The idea
behind UCB is to pick actions that maximise
\(\mathbb{E[R]} + \kappa \sigma[R]\), where \(\sigma[R]\) is the
standard deviation of the return and \(\kappa > 0\). UCB therefore
encourages exploration in regions with high uncertainty and moderate
expected return. Whilst easily achievable in small tabular cases, the
use of powerful density models \citep{bellemare2016unifying}, or
conversely, hashing \citep{tang2017exploration}, has allowed this
algorithm to scale to high-dimensional visual domains with DRL. UCB is
only one technique for trading off exploration and exploitation in the
context of Bayesian optimisation \citep{shahriari2016taking}; future
work in DRL may benefit from investigating other successful techniques
that are used in Bayesian optimisation.

UCB can also be considered one way of implementing \emph{intrinsic
motivation}, which is a general concept that advocates decreasing
uncertainty\slash making progress in learning about the environment
\citep{schmidhuber1991possibility}. There have been several DRL
algorithms that try to implement intrinsic motivation via minimising
model prediction error
\citep{stadie2015incentivizing, pathak2017curiosity} or maximising
information gain \citep{mohamed2015variational, houthooft2016vime}.

\subsection{Hierarchical RL}\label{hierarchical-rl}

In the same way that deep learning relies on hierarchies of features,
HRL relies on hierarchies of policies. Early work in this area
introduced \emph{options}, in which, apart from \emph{primitive actions}
(single-timestep actions), policies could also run other policies
(multi-timestep ``actions'') \citep{sutton1999between}. This approach
allows top-level policies to focus on higher-level \emph{goals}, whilst
\emph{subpolicies} are responsible for fine control. Several works in
DRL have attempted HRL by using one top-level policy that chooses
between subpolicies, where the division of states or goals in to
subpolicies is achieved either manually
\citep{arulkumaran2016classifying, tessler2017deep, kulkarni2016hierarchical}
or automatically
\citep{bacon2017option, vezhnevets2016strategic, vezhnevets2017feudal}.
One way to help construct subpolicies is to focus on discovering and
reaching goals, which are specific states in the environment; they may
often be locations, which an agent should navigate to. Whether utilised
with HRL or not, the discovery and generalisation of goals is also an
important area of ongoing research
\citep{schaul2015universal, kulkarni2016deep, vezhnevets2017feudal}.

\subsection{Imitation Learning and Inverse
RL}\label{imitation-learning-and-inverse-rl}

One may ask why, if given a sequence of ``optimal'' actions from expert
demonstrations, it is not possible to use supervised learning in a
straightforward manner---a case of ``learning from demonstration''. This
is indeed possible, and is known as \emph{behavioural cloning} in
traditional RL literature. Taking advantage of the stronger signals
available in supervised learning problems, behavioural cloning enjoyed
success in earlier neural network research, with the most notable
success being ALVINN, one of the earliest autonomous cars
\citep{pomerleau1989alvinn}. However, behavioural cloning cannot adapt
to new situations, and small deviations from the demonstration during
the execution of the learned policy can compound and lead to scenarios
where the policy is unable to recover. A more generalisable solution is
to use provided trajectories to guide the learning of suitable
state-action pairs, but fine-tune the agent using RL
\citep{hester2017learning}. Alternatively, if the expert is still
available to query during training, the agent can use active learning to
gather extra data when it is unsure, allowing it to learn from states
away from the optimal trajectories \citep{ross2011reduction}. This has
been applied to a deep learning setting, where a CNN trained in a visual
navigation task with active learning significantly improved upon a pure
imitation learning baseline \citep{hussein2016deep}.

The goal of IRL is to estimate an unknown reward function from observed
trajectories that characterise a desired solution
\citep{ng2000algorithms}; IRL can be used in combination with RL to
improve upon demonstrated behaviour. Using the power of deep neural
networks, it is now possible to learn complex, nonlinear reward
functions for IRL \citep{wulfmeier2015maximum}. Ho and Ermon
\citep{ho2016generative} showed that policies are uniquely characterised
by their \emph{occupancies} (visited state and action distributions)
allowing IRL to be reduced to the problem of measure matching. With this
insight, they were able to use generative adversarial training
\citep{goodfellow2014generative} to facilitate reward function learning
in a more flexible manner, resulting in the generative adversarial
imitation learning (GAIL) algorithm. GAIL was later extended to allow
IRL to be applied even when receiving expert trajectories from a
different visual viewpoint to that of the RL agent
\citep{stadie2017third}. In complementary work, Baram et al.
\citep{baram2016model} exploit gradient information that was not used in
GAIL to learn models within the IRL process.

\subsection{Multi-agent RL}\label{multi-agent-rl}

Usually, RL considers a single learning agent in a stationary
environment. In contrast, multi-agent RL (MARL) considers multiple
agents learning through RL, and often the non-stationarity introduced by
other agents changing their behaviours as they learn
\citep{busoniu2008comprehensive}. In DRL, the focus has been on enabling
(differentiable) communication between agents, which allows them to
co-operate. Several approaches have been proposed for this purpose,
including passing messages to agents sequentially
\citep{foerster2016learning}, using a bidirectional channel (providing
ordering with less signal loss) \citep{peng2017multiagent}, and an
all-to-all channel \citep{sukhbaatar2016learning}. The addition of
communication channels is a natural strategy to apply to MARL in complex
scenarios and does not preclude the usual practice of modelling
co-operative or competing agents as applied elsewhere in the MARL
literature \citep{busoniu2008comprehensive}. Other DRL works of note in
MARL investigate the effects of learning and sequential decision making
in game theory \citep{heinrich2016deep, leibo2017multi}.

\subsection{Memory and Attention}\label{memory-and-attention}

As one of the earliest works in DRL the DQN spawned many extensions. One
of the first extensions was converting the DQN into an RNN, which allows
the network to better deal with POMDPs by integrating information over
long time periods. Like recursive filters, recurrent connections provide
an efficient means of acting conditionally on temporally distant prior
observations. By using recurrent connections between its hidden units,
the deep recurrent \(Q\)-network (DRQN) introduced by Hausknecht and
Stone \citep{hausknecht2015deep} was able to successfully infer the
velocity of the ball in the game ``Pong,'' even when frames of the game
were randomly blanked out. Further improvements were gained by
introducing \emph{attention}---a technique where additional connections
are added from the recurrent units to lower layers---to the DRQN,
resulting in the deep attention recurrent \(Q\)-network (DARQN)
\citep{sorokin2015deep}. Attention gives a network the ability to choose
which part of its next input to focus on, and allowed the DARQN to beat
both the DQN and DRQN on games, which require longer-term planning.
However, the DQN outperformed the DRQN and DARQN on games requiring
quick reactions, where \(Q\)-values can fluctuate more rapidly.

Taking recurrent processing further, it is possible to add a
differentiable memory to the DQN, which allows it to more flexibly
process information in its ``working memory'' \citep{oh2016control}. In
traditional RNNs, recurrent units are responsible for both performing
calculations and storing information. Differentiable memories add large
matrices that are purely used for storing information, and can be
accessed using differentiable read and write operations, analagously to
computer memory. With their key-value-based memory \(Q\)-network (MQN),
Oh et al. \citep{oh2016control} constructed an agent that could solve a
simple maze built in Minecraft, where the correct goal in each episode
was indicated by a coloured block shown near the start of the maze. The
MQN, and especially its more sophisticated variants, significantly
outperformed both DQN and DRQN baselines, highlighting the importance of
using decoupled memory storage. More recent work, where the memory was
given a 2D structure in order to resemble a spatial map, hints at future
research where more specialised memory structures will be developed to
address specific problems, such as 2D or 3D navigation
\citep{parisotto2017neural}. Alternatively, differentiable memories can
be used as approximate hash tables, allowing DRL algorithms to store and
retrieve successful experiences to facilitate rapid learning
\citep{pritzel2017neural}.

Note that RNNs are not restricted to value-function-based methods but
have also been successfully applied to policy search
\citep{wierstra2010recurrent} and actor-critic methods
\citep{heess2015memory, mnih2016asynchronous}.

\subsection{Transfer Learning}\label{transfer-learning}

Even though DRL algorithms can process high-dimensional inputs, it is
rarely feasible to train RL agents directly on visual inputs in the real
world, due to the large number of samples required. To speed up learning
in DRL, it is possible to exploit previously acquired knowledge from
related tasks, which comes in several guises: transfer learning,
multitask learning \citep{caruana1997multitask} and curriculum learning
\citep{bengio2009curriculum} to name a few. There is much interest in
transferring learning from one task to another, particularly from
training in physics simulators with visual renderers and fine-tuning the
models in the real world. This can be achieved in a naive fashion,
directly using the same network in both the simulated and real phases
\citep{zhu2017target}, or with more sophisticated training procedures
that directly try to mitigate the problem of neural networks
``catastrophically forgetting'' old knowledge by adding extra layers
when transferring domain \citep{rusu2016progressive, rusu2017sim}. Other
approaches involve directly learning an alignment between simulated and
real visuals \citep{tzeng2016towards}, or even between two different
camera viewpoints \citep{stadie2017third}.

A different form of transfer can be utilised to help RL in the form of
multitask training
\citep{li2015recurrent, jaderberg2017reinforcement, mirowski2017learning}.
Especially with neural networks, supervised and unsupervised learning
tasks can help train features that can be used by RL agents, making
optimising the RL objective easier to achieve. For example, the
``unsupervised reinforcement and auxiliary learning'' A3C-based agent is
additionally trained with ``pixel control'' (maximally changing pixel
inputs), plus reward prediction and value function learning from
experience replay \citep{jaderberg2017reinforcement}. Meanwhile, the
A3C-based agent of Mirowski et al. \citep{mirowski2017learning} was
additionally trained to construct a depth map given RGB inputs, which
helps it in its task of learning to navigate a 3D environment. In an
ablation study, Mirowski et al. \citep{mirowski2017learning} showed the
predicting depth was more useful than receiving depth as an extra input,
lending further support to the idea that gradients induced by auxiliary
tasks can be extremely effective at boosting DRL.

Transfer learning can also be used to construct more data- and
parameter-efficient policies. In the student-teacher paradigm in machine
learning, one can first train a more powerful ``teacher'' model, and
then use it to guide the training of a less powerful ``student'' model.
Whilst originally applied to supervised learning, the neural network
knowledge transfer technique known as \emph{distillation}
\citep{hinton2014distilling} has been utilised to both transfer policies
learned by large DQNs to smaller DQNs, and transfer policies learned by
several DQNs trained on separate games to one single DQN
\citep{parisotto2016actor, rusu2016policy}. Together, the combination of
multitask and transfer learning can improve the sample efficiency and
robustness of current DRL algorithms \citep{teh2017distral}. These are
important topics if we wish to construct agents that can accomplish a
wide range of tasks, since naively training on multiple RL objectives at
once may be infeasible.

\subsection{Benchmarks}\label{benchmarks}

One of the challenges in any field in machine learning is developing a
standardised way to evaluate new techniques. Although much early work
focused on simple, custom MDPs, there shortly emerged control problems
that could be used as standard benchmarks for testing new algorithms,
such as the Cartpole \citep{barto1983neuronlike} and Mountain Car
\citep{moore1990efficient} domains.

However, these problems were limited to relatively small state spaces,
and therefore failed to capture the complexities that would be
encountered in most realistic scenarios. Arguably the initial driver of
DRL, the ALE provided an interface to Atari 2600 video games, with code
to access over 50 games provided with the initial release
\citep{bellemare2015arcade}. As video games can vary greatly, but still
present interesting and challenging objectives for humans, they provide
an excellent testbed for RL agents. As the first algorithm to
successfully play a range of these games directly from their visuals,
the DQN \citep{mnih2015human} has secured its place as a milestone in
the development of RL algorithms. This success story has started a trend
of using video games as standardised RL testbeds, with several
interesting options now available. ViZDoom provides an interface to the
Doom first-person shooter \citep{kempka2016vizdoom}, and echoing the
popularity of e-sports competitions, ViZDoom competitions are now held
at the yearly IEEE Conference on Computational Intelligence in Games.
Facebook's TorchCraft \citep{synnaeve2016torchcraft} and DeepMind's
StarCraft II Learning Environment \citep{vinyals2017starcraft}
respectively provide interfaces to the StarCraft and StarCraft II
real-time strategy games, presenting challenges in both micromanagement
and long-term planning. In an aim to provide more flexible environments,
DeepMind Lab was developed on top of the Quake III Arena first-person
shooter engine \citep{beattie2016deepmind}, and Microsoft's Project
Malmo exposed an interface to the Minecraft sandbox game
\citep{johnson2016malmo}. Both environments provide customisable
platforms for RL agents in 3D environments.

Most DRL approaches focus on discrete actions, but some solutions have
also been developed for continuous control problems. Many DRL papers in
continuous control
\citep{schulman2015trust, heess2015learning, lillicrap2016continuous, mnih2016asynchronous, baram2016model, stadie2017third}
have used the MuJoCo physics engine to obtain relatively realistic
dynamics for multi-joint continuous control problems
\citep{todorov2012mujoco}, and there has now been some effort to
standardise these problems \citep{duan2016benchmarking}.

To help with standardisation and reproducibility, most of the
aforementioned RL domains and more have been made available in the
OpenAI Gym, a library and online service that allows people to easily
interface with and publicly share the results of RL algorithms on these
domains \citep{brockman2016openai}.

\section{Conclusion: Beyond Pattern
Recognition}\label{conclusion-beyond-pattern-recognition}

Despite the successes of DRL, many problems need to be addressed before
these techniques can be applied to a wide range of complex real-world
problems \citep{lake2016building}. Recent work with (non-deep)
generative causal models demonstrated superior generalisation over
standard DRL algorithms
\citep{mnih2016asynchronous, rusu2016progressive} in some benchmarks
\citep{bellemare2015arcade}, achieved by reasoning about causes and
effects in the environment \citep{kansky2017schema}. For example, the
schema networks of Kanksy et al. \citep{kansky2017schema} trained on the
game ``Breakout'' immediately adapted to a variant where a small wall
was placed in front of the target blocks, whilst progressive (A3C)
networks \citep{rusu2016progressive} failed to match the performance of
the schema networks even after training on the new domain. Although DRL
has already been combined with AI techniques, such as search
\citep{silver2016mastering} and planning \citep{tamar2016value}, a
deeper integration with other traditional AI approaches promises
benefits such as better sample complexity, generalisation and
interpretability \citep{garnelo2016towards}. In time, we also hope that
our theoretical understanding of the properties of neural networks
(particularly within DRL) will improve, as it currently lags far behind
practice.

To conclude, it is worth revisiting the overarching goal of all of this
research: the creation of general-purpose AI systems that can interact
with and learn from the world around them. Interaction with the
environment is simultaneously the advantage and disadvantage of RL.
Whilst there are many challenges in seeking to understand our complex
and ever-changing world, RL allows us to choose how we explore it. In
effect, RL endows agents with the ability to perform experiments to
better understand their surroundings, enabling them to learn even
high-level causal relationships. The availability of high-quality visual
renderers and physics engines now enables us to take steps in this
direction, with works that try to learn intuitive models of physics in
visual environments \citep{denil2017learning}. Challenges remain before
this will be possible in the real world, but steady progress is being
made in agents that learn the fundamental principles of the world
through observation and action. Perhaps, then, we are not too far away
from AI systems that learn and act in more human-like ways in
increasingly complex environments.

\section*{Acknowledgments}

The authors would like to thank the reviewers and broader community for
their feedback on this survey; in particular, we would like to thank
Nicolas Heess for clarifications on several points. Kai Arulkumaran
would like to acknowledge PhD funding from the Department of
Bioengineering, Imperial College London. This research has been
partially funded by a Google Faculty Research Award to Marc Deisenroth.

%
%


\ifCLASSOPTIONcaptionsoff
  \newpage
\fi



%

\bibliography{references.bib}


%

  \begin{IEEEbiographynophoto}{Kai Arulkumaran} (ka709@imperial.ac.uk) is a Ph.D.~candidate in the Department of
Bioengineering at Imperial College London. He received a B.A. in
Computer Science at the University of Cambridge in 2012, and an M.Sc. in
Biomedical Engineering at Imperial College London in 2014. He was a
Research Intern in Twitter Magic Pony and Microsoft Research in 2017.
His research focus is deep reinforcement learning and transfer learning
for visuomotor control.
  \end{IEEEbiographynophoto}
  \begin{IEEEbiographynophoto}{Marc Peter Deisenroth} (m.deisenroth@imperial.ac.uk) is a Lecturer in Statistical Machine
Learning in the Department of Computing at Imperial College London and
with PROWLER.io. He received an M.Eng. in Computer Science at the
University of Karlsruhe in 2006 and a Ph.D.~in Machine Learning at the
Karlsruhe Institute of Technology in 2009. He has been awarded an
Imperial College Research Fellowship in 2014 and received Best Paper
Awards at ICRA 2014 and ICCAS 2016. He is a recipient of a Google
Faculty Research Award and a Microsoft Ph.D.~Scholarship. His research
is centred around data-efficient machine learning for autonomous
decision making.
  \end{IEEEbiographynophoto}
  \begin{IEEEbiographynophoto}{Miles Brundage} (miles.brundage@philosophy.ox.ac.uk) is a Ph.D.~candidate in Human and
Social Dimensions of Science and Technology at Arizona State University,
and a Research Fellow at the University of Oxford's Future of Humanity
Institute. He received a B.A. in Political Science at George Washington
University in 2010. His research focuses on governance issues related to
artificial intelligence.
  \end{IEEEbiographynophoto}
  \begin{IEEEbiographynophoto}{Anil Anthony Bharath} (a.bharath@imperial.ac.uk) is a Reader in the Department of
Bioengineering at Imperial College London and a Fellow of the
Institution of Engineering and Technology. He received a B.Eng. in
Electronic and Electrical Engineering from University College London in
1988, and a Ph.D.~in Signal Processing from Imperial College London in
1993. He was an academic visitor in the Signal Processing Group at the
University of Cambridge in 2006. He is a co-founder of Cortexica Vision
Systems. His research interests are in deep architectures for visual
inference.
  \end{IEEEbiographynophoto}





\end{document}